\documentclass[acmsmall,screen,authorversion,nonacm]{acmart}

\AtBeginDocument{%
  }

\usepackage{wrapfig}
\usepackage{amsthm} %

\usepackage{xspace}
\newcommand{\approach}[0]{\textsc{codealign}\xspace}

\usepackage{subcaption}
\usepackage{booktabs}
\usepackage{cancel} %
\usepackage{multirow}
\usepackage{hyperref}
\usepackage{algorithm2e}
\usepackage{amsmath}
  
\usepackage{listings}
\definecolor{dkgreen}{rgb}{0,0.5,0}
\definecolor{lessdkgreen}{rgb}{0,0.6,0}
\definecolor{dkred}{rgb}{0.5,0,0}
\definecolor{gray}{rgb}{0.5,0.5,0.5}

\lstdefinestyle{cstyle}{
language=c,
basicstyle=\ttfamily\bfseries\scriptsize,
  morekeywords={virtualinvoke},
  keywordstyle=\color{blue},
  ndkeywordstyle=\color{red},
  commentstyle=\color{dkred},
  stringstyle=\color{dkgreen},
  numbers=left,
  breaklines=true,
  numberstyle=\ttfamily\footnotesize\color{gray},
  stepnumber=1,
  numbersep=2pt,
  numberstyle=\tiny,
  backgroundcolor=\color{white},
  tabsize=4,
  showspaces=false,
  showstringspaces=false,
  xleftmargin=.23in,
  captionpos=b,
  escapeinside={$}{$},
  print
}
\lstdefinestyle{cinlinestyle}{
language=c,
basicstyle=\ttfamily\bfseries\footnotesize,
  morekeywords={virtualinvoke},
  keywordstyle=\color{blue},
  ndkeywordstyle=\color{red},
  commentstyle=\color{dkred},
  stringstyle=\color{dkgreen},
  escapeinside={$}{$},
  print
}

\newcommand\cinline[1]{{\lstinline[style=cinlinestyle]@#1@}}

\begin{document}

\title{Fast, Fine-Grained Equivalence Checking for Neural Decompilers}

\author{Luke Dramko}
\email{lukedram@cs.cmu.edu}
\author{Claire Le Goues}
\email{clegoues@cs.cmu.edu}
\affiliation{%
  \institution{Carnegie Mellon University}
  \city{Pittsburgh}
  \state{PA}
  \country{USA}
}

\author{Edward J. Schwartz}
\affiliation{%
  \institution{Carnegie Mellon University Software Engineering Institute}
  \city{Pittsburgh}
  \state{PA}
  \country{USA}}
\email{eschwartz@cert.org}

\renewcommand{\shortauthors}{Dramko et al.}

\begin{abstract}
Neural decompilers are machine learning models that reconstruct the source code from an executable program.
Critical to the lifecycle of any machine learning model is an evaluation of its effectiveness.
However, existing techniques for evaluating neural decompilation models have substantial weaknesses, especially when it comes to showing the correctness of the neural decompiler's predictions.
To address this, we introduce \approach, a novel instruction-level code equivalence technique designed for neural decompilers.
We provide a formal definition of a relation between equivalent instructions, which we term an equivalence alignment.
We show how \approach generates equivalence alignments, then evaluate \approach by comparing it with symbolic execution.
Finally, we show how the information \approach provides---which parts of the functions are equivalent and how well the variable names match---is substantially more detailed than existing state-of-the-art evaluation metrics, which report unitless numbers measuring similarity.
\end{abstract}

\ccsdesc[500]{Software and its engineering~Functionality}
\ccsdesc[100]{Software and its engineering~Software maintenance tools}

\keywords{Program Equivalence, Alignment, Program Analysis}

\received{20 February 2007}
\received[revised]{12 March 2009}
\received[accepted]{5 June 2009}

\maketitle

\section{Introduction}

Native decompilation is the process of reconstructing source code from a compiled executable.
Decompilation is used for several security-related code maintenance tasks, including malware analysis, vulnerability research, and patching legacy software for which the corresponding source code is not available.
Because a significant amount of information is discarded during compilation, including variables and their names and types, decompilation cannot fully be solved deterministically.
Deterministic conventional decompilers focus on code semantics and are constructed using architectures similar to an optimizing compiler~\cite{vanemmerik:2007}, and while an improvement on machine code, still produce difficult-to-read reconstructed programs, because they do not attempt to recover elements like meaningful variable names and types.

As a result, a recent surge of interest has applied neural learning to either recover a specific feature such as variable names~\cite{dire,dirty,varbert,hext5} or to decompile an entire program~\cite{Katz2018,coda,wu2022dnd,cao2022boosting,liang2021neutron,hudegpt,nova,slade,llm4decompile}.
The latter class of models are termed \emph{neural decompilers}.
Neural decompilers are probabilistic, sampling from a distribution of possible source code representations. 
In theory, they should be able to reconstruct the original source code in some cases.
However, they can also hallucinate, and produce code that is not semantically equivalent to the original.
Determining how often hallucinations occur is critical for knowing how much a given neural decompiler can be trusted.

To perform such evaluations, researchers usually set up an experiment where the correct answer---i.e., the original source code---is available; the neural decompiler's predictions are compared against the original.
However, program equivalence is an extremely difficult problem that is undecidable in general.
Researchers have created a variety of domain-specific program equivalence techniques that meet the needs of particular use cases, balancing trade-offs of soundness, completeness, efficiency, and applicability.
For example, formal, sound, but expensive and incomplete methodologies are used to prove that code produced by an optimizing compiler is equivalent to the original~\cite{gupta2018effective,churchill2019semantic}.
On the other hand, complete, widely applicable, and fast but very unsound methods can be employed to approximate equivalence in a loose way when evaluating machine learning models that generate code against a reference code snippet~\cite{codebleu,codebertscore,crystalbleu,bleu,tran2019does}.
They use a variety of heuristics, such as syntactic or the lexical overlap between functions, to produce a score that reflects how similar the two functions are.

\begin{figure*}[!t]
\includegraphics[width=\textwidth]{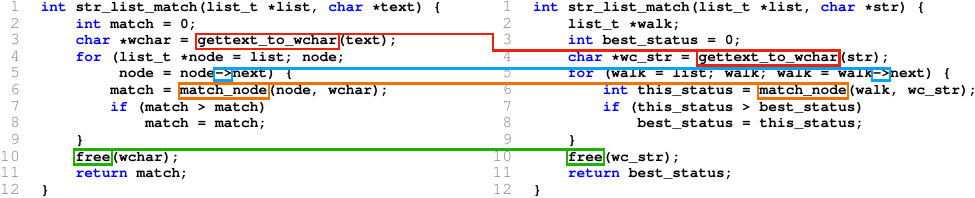}
\caption{\small A prediction by a machine learning model (left) and the reference solution (right).
Some instructions in the prediction are equivalent to the reference;
these are connected with boxes and lines.
However, the prediction is subtly incorrect; it always returns the value of evaluating \cinline{match_node} on the \emph{last} item of the list, rather than the highest one found.
Equivalent instructions are connected with boxes and lines.}\label{fig:introexample}
\end{figure*}

Beyond equivalence of entire functions, it is often desirable to know what \emph{parts} of the neural decompiler's predictions are equivalent to the reference.
We illustrate with an example.
Fig.~\ref{fig:introexample} shows such a pair of functions, the left is generated a machine learning model, while the right is the correct answer.
The non-control-flow instructions that are equivalent are connected with boxes and lines.
The reference code searches through a linked list to find the maximum value.
The generated code correctly manages memory for a temporary object (the \cinline{gettext_to_wchar} and \cinline{free} calls are equivalent) and correctly interates through the list (the \cinline{->} operations are equivalent), albeit with slightly different syntax and variable names.
However, the model had a hallucination: it uses just a single variable to represent both of the current and maximum value.
The \cinline{match > match} and \cinline{match = match} are not equivalent to \cinline{this_status > best_status} and \cinline{best_status = this_status}.
As a result, the generated function always returns the value of the last element in the list.

In this work, we introduce \approach, an instruction-level equivalence technique designed for neural decompilation.
\approach produces an \emph{equivalence alignment}: a relation of equivalent instructions between two functions.
Intuitively, two instructions are equivalent if and only if the result of executing those instructions is the same for all inputs.
(For a formal definition see Section~\ref{sec:definitions}).
The boxes and lines in Fig.~\ref{fig:introexample} illustrate an equivalence alignment.
This granular, low-level equivalence representation allows for detailed analyses of the results.

Evaluating neural decompilers offers unique challenges that break assumptions made by other code equivalence techniques.
Ironically, code produced by decompilers, including neural decompilers, usually can't be compiled.
This is because external symbols that compilers assume exist, like functions and global variables, are unavailable.
Further, even if the code could be compiled, there are rarely any test suites available in situations where decompilation is used.
(Tests can be used as a proxy for equivalence themselves or for performing trace alignment as~\cite{churchill2019semantic} do).
Generating tests automatically is theoretically possible but extremely difficult in practice, especially when dealing with sophisticated data structures that occur in real code, many of which have invariants over themselves (e.g. a binary search tree is a tree structure with an ordering invariant) or which depend on state external to the program (e.g. an open socket).
In addition, in malware analysis, an important use case for decompilation, executing pieces of the program can be dangerous, because that means executing the malware.
Finally, evaluating a machine learning model often involves processing hundreds or thousands of predictions, so runtime performance is important.

Fortunately, the nature of the neural decompilation affords opportunities as well.
A key insight that enables \approach is that the decompilation and original code should be \emph{implementations of the same algorithm}.
Compilation is a lossy, many-to-one function: textually distinct source code can map to the same sequence of assembly instructions, especially with optimizations.
A neural decompiler therefore cannot deterministically match the original source code all of the time.
Frequently, there will be differences in variable name and (irrelevant) statement ordering; expressions may also be broken up differently.
However, the assembly instructions necessarily preserve the algorithm itself accuracy, and so a neural decompiler should be able to reproduce the algorithm in some form.

Traditional approaches for evaluating code predicted by machine learning, such as CodeBLEU~\cite{codebleu}, and CodeBERTScore~\cite{codebertscore}, use various heuristics to approximate program equivalence; for Fig.~\ref{fig:introexample}, their values are 0.595 and 0.905, respectively.
However, it is not clear how to interpret these scores; they are unitless.
(These numbers should not be interpreted as proportions).
In particular, it is not clear how these values describe the partially equivalent behavior of Fig.~\ref{fig:introexample}.
Nor is it the case that a sufficiently high score indicates equivalence, as we show in Section~\ref{sec:neuraldecompilation}.

Identifying which parts of a function are equivalent can also help measure different aspects of similarity in isolation, such as functional correctness and variable name similarity.
For example, to measure how well the variable names in generated code match those in the reference, one can identify the correspondence between variables in the generated code and variables in the reference code.
With this correspondence, the generated variable names can be compared to the reference names. %
For instance, in Fig.~\ref{fig:introexample}, \cinline{walk} and \cinline{node} are equivalent; they both store the current node while iterating over the list.  However, their names are not similar, which could be confusing.
\approach's equivalence alignments make it easy to determine which variables correspond.

We contribute the following:
\begin{itemize}
\item the \approach tool which builds alignments from C (and has some support for Python)
\item an equivalence alignment generator based on symbolic execution for comparison
\item a demonstration of \approach's utility in evaluating both the correctness and variable name quality of code produced by a neural decompiler against a reference
\end{itemize}

\section{Definitions}
\label{sec:definitions}

\approach operates on pairs of functions.
We model each function as a sequence of instructions $p_j$.
Each instruction operates on values $v_i$, where each value is either an argument to the function, a constant, or the result of executing another instruction.  These values are called the instruction's \emph{operands}.
The class of computation that an instruction performs is defined by its \emph{operator}, such as \cinline{+}, \cinline{>}, and \cinline{strcmp}.
As an example from Fig.~\ref{fig:introexample}, \cinline{this_status > best_status} is an instruction; \cinline{this_status} and \cinline{best_status} are its operands; and \cinline{>} is its operator.
All of the items in boxes in Fig.~\ref{fig:introexample} are operators.

\paragraph{Value-instruction Binding} 
Without loss of generality, we define an equivalence alignment in terms of SSA-form code~\cite{rosen1988ssa,alpern1988ssa}.
In SSA form, variables are statically immutable: they are assigned to at most once during the program.
We borrow the term \emph{value} from the SSA literature (and in particular the LLVM compiler infrastructure) to refer to any object that can function as the operand to an instruction.
This includes SSA-style immutable static variables, but also constants and the enclosing function's arguments.

A value binding is equivalent to the assignment of an instruction to an SSA immutable static variable.
Let $f$ be a function, let $v \in f$ be a value, and let $p_j$ be an instruction. We say that $v_j$ is bound to $p_j$ if $v_j$ is the result of executing $p_j$.

\paragraph{Equivalent Values}
A common way to define functions as semantically equivalent is to say that two functions are equivalent if they have the same output for every input.
Formally, $ f = g \iff \forall i, f(i) = g(i)$.
We use a similar definition to define equivalent values.
Let $v$ be a value and let $v^{f(i)}$ be the \emph{dynamic} value of $v$ when function $f$ is executed on input $i$.
Then
\begin{equation}
v_j = v_k \iff \forall i, v_j^{f(i)} = v_k^{g(i)}.
\label{eq:functionalequivalence}
\end{equation}
We say that $v_j$ and $v_k$ are \emph{functionally equivalent}.
Of course, a given static value $v$ may have multiple dynamic values if that variable occurs within a loop.
To define the equivalence of values in a loop, we use induction.
Any variable that is changed within a loop has an associated $\phi$ instruction at its head.
Such a $\phi$ instruction is equivalent to another if:
\begin{itemize}
\item \emph{Base case}: the values of the $\phi$ instructions' operands from outside the loop are equivalent. (There is always such an operand because all loops have an entry point).
\item \emph{Inductive Step}: given equivalent $\phi$ instructions, the values of their operands inside the loop are equivalent.
\end{itemize}

\paragraph{Equivalence Alignments} 
An equivalence alignment is a relation between instructions in two functions.
Let $f$ and $g$ be two functions, and let $F$ and $G$ be sets of all values that occur in each function, respectively.
Let $v_j$ be the instruction bound to value $p_j$.
We define an equivalence alignment as a subset of the cartesian product $F \times G$:
\begin{equation}
\label{eq:alignment}
\{(p_j, p_k) | v_j \in F, v_k \in G, v_j = v_k\}
\end{equation}
That is, an equivalence alignment consists of the instructions whose results are equivalent.
An equivalence alignment is a relation; it is not necessarily a function.

Note that this definition of an equivalence alignment, is, like program equivalence in general, undecidable.
In practice, \approach uses a sound but incomplete definition of equivalence, which we discuss in Section~\ref{sec:lemmageneration}.
An equivalence alignment is related to a product program~\cite{barthe2011relational}, where elements related to each other are functionally equivalent.

\section{Codealign}
\label{sec:approach}

\begin{figure*}
\begin{subfigure}{\textwidth}
\noindent\begin{minipage}{0.37\textwidth}
\begin{lstlisting}[style=cstyle,basicstyle=\ttfamily\bfseries\tiny]
void f(char *str, size_t len) {
  for (size_t i = 0; i < len; i++)
    write(1, writable(str + i), 1);
}
\end{lstlisting}
\vspace{48pt}
\end{minipage}
\begin{minipage}{0.37\textwidth}
\begin{lstlisting}[style=cstyle,basicstyle=\ttfamily\bfseries\tiny]
void g(char *ptr, size_t size) {
  size_t i;
  i = 0;
  while (i < size) {
    write(1, writable(ptr + i), 1);
    i++;
  }
}
\end{lstlisting}
\vspace{16pt}
\end{minipage}
\begin{minipage}{0.25\textwidth}
\begin{lstlisting}[style=cstyle,basicstyle=\ttfamily\bfseries\tiny,mathescape=true]
ssa_f(str, len) {
  loop (%
    write(1, %
    }
}
\end{lstlisting}
\end{minipage}
\caption{\small Two functions, $f$ (left), $g$ (middle) and $f$'s SSA form (right). The SSA form of $g$ is the same as $f$'s except for parameter names. Despite their different syntax, each value in each function is equivalent to one value in the other. Because $f$ and $g$ contain multiple $+$ instructions, we use the source line to differentiate them, e.g., $+_{(3),f}$ refers to the $+$ on line 3 in $f$. We adopt LLVM's convention of illustrating SSA instruction values as \%0, \%1, etc.}
\label{fig:inference_a}
\end{subfigure}
\vspace{4pt}

\begin{subfigure}{\textwidth}
\includegraphics[width=1.0\textwidth]{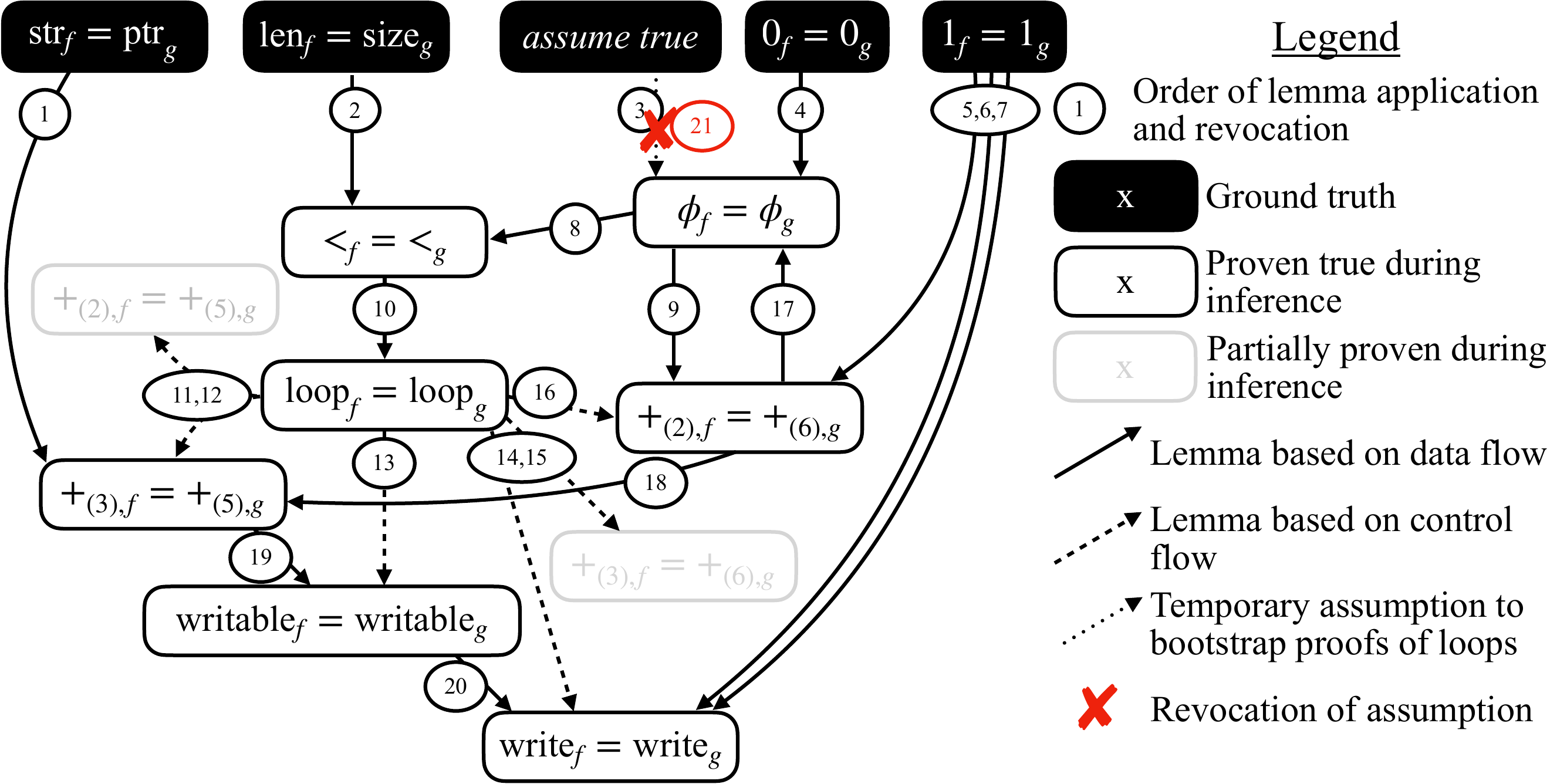}
\caption{\small An inductive proof graph showing values from the functions in Fig.~\ref{fig:inference_a} equivalent.  An instruction in $f$ can be proven equivalent to an instruction in $g$ if all of its' control- and dataflow dependencies can be proven equivalent. The graph is initialized with only ground truth nodes, and adds others as lemmas are applied in the order specified.}
\label{fig:inference_d}
\end{subfigure}

\caption{\small An illustration of how \approach works. Fig.~\ref{fig:inference_a} shows two examples (derived from the data in Section~\ref{sec:neuraldecompilation}) and their canonicalized SSA representations. Fig.~\ref{fig:inference_d} shows how, using pairs of SSA-representation edges as lemmas, various values in the examples can be proven equivalent.}
\label{fig:inference}
\end{figure*}

\approach takes two functions $f$ and $g$ as input and outputs an equivalence alignment.  
Fig.~\ref{fig:inference} shows an example pair of functions $f$ and $g$ and their internal representations as they progress through the different stages of \approach.
To begin, \approach converts each function to SSA form, and computes control dependencies for each instruction.
Using this information, it creates lemmas that will be used to prove parts of $f$ and $g$ equivalent.
In the final step, \approach iteratively proves that instructions are equivalent using induction (Fig.~\ref{fig:inference_d}).%

\subsection{Pre-processing}

Because \approach operates on functions in isolation, and does not require header files or libraries, it first heuristically classifies unknown identifiers in functions based on their use.
For example, undeclared identifiers in call expressions are assumed to represent functions.
If the code treats an undeclared identifier as a value, \approach interprets it as a global variable.
\approach then converts the IR to SSA form, performs standard copy propagation, and applies normalization and desugaring rules (e.g., converting \cinline{x++} to \cinline{x = x + 1}) designed to enable very similar code to be detected as equivalent. Fig.~\ref{fig:inference_a} shows an example.
 
\approach operates on an SSA-based data flow graph: nodes represent instructions, and edges represent a data-flow between them.
Note that incoming edges, which represent the operands of a given instruction, are \emph{ordered} because the order of operands is semantically important.  The graph also treats SSA $\phi$ instructions as instructions that receive their own nodes.
\approach also uses control dependence information from a control dependence graph; nodes in the control dependence graph represent basic blocks and edges represent the control dependencies between basic blocks.

\subsection{Lemma Generation}
\label{sec:lemmageneration}

Pairs of equivalent values, one from $f$ and the other from $g$, serve as propositions in \approach's logical system.
Lemmas are implications that relate information about propositions to each other.
\approach uses the SSA representations of the functions as well as control dependence information to generate lemmas to be used in the inductive phase.

Because finding all functionally equivalent values, as defined in Section~\ref{sec:definitions}, is undecidable,
\approach uses a sound but incomplete notion of equivalence, which we term \emph{dependency-based equivalence}.
Dependency-based equivalence is based on the intuition that performing the same computation (that is, executing instructions with the same operator) on the same operands under the same control flow conditions will produce the same result.
Let $v_j\in f$ and $v_k\in g$ be values bound to instructions. 
Informally, lemmas take the form
\begin{equation}
\text{all dependencies of } v_j \text{ and } v_k \text{ align } \wedge \text{operator}(v_j) = \text{operator}(v_k) \implies v_j = v_k
\label{eq:biglemma}
\end{equation}
Naively, there could be an equivalence lemma for each combination of instructions from $f$ and $g$.
If there are $n$ instructions in $f$ and $m$ instructions in $g$, then there would be $nm$ lemmas.
However, if the operators for two instructions are different then it is impossible to satisfy the conditions of any lemma of the form given in Equation~\ref{eq:biglemma}.
Thus, \approach simply does not generate these lemmas, which speeds up lemma generation.

In \approach, we use an equivalent but slightly different form for representing lemmas that makes it easier to determine when the conditions of a lemma are satisfied.
Instead of generating one lemma for each pair of instructions with $n$ dependencies $\{d_1,\dots d_i,\dots d_n\}$, we generate $n$ lemmas, one for each pair of dependencies.
(In \approach, incoming edges are ordered, so there are not $n^2$; we discuss ordering below).
The conclusion of each lemma is associated with a weight of $1/n$; the conclusion is only considered proven if the total proven weight is 1.
These take the form
\begin{equation}
d_{i,j} = d_{i,k} \implies (v_{j} = v_{k}, \text{weight}=1/n)
\label{eq:smalllemma}
\end{equation}
The weight formulation is useful for how we handle loop induction (Section~\ref{sec:loops}).
In Fig.~\ref{fig:inference}, there are three lemmas concluding in $+_{(3),f}=+_{(5),g}$, where $+_{(3),f}$ refers to the \cinline{+} on line 3 of $f$ and $+_{(5),g}$ refers to the \cinline{+} on line 5 of $g$. 
\begin{enumerate}
\item $\text{str}_f = \text{ptr}_g \implies +_{(3),f}=+_{(5),g}$
\item $+_{(2),f} = +_{(6),g}  \implies +_{(3),f}=+_{(5),g}$
\item $\text{loop}_f=\text{loop}_g \implies +_{(3),f}=+_{(5),g}$
\end{enumerate}
The first two are data flow lemmas, and the third is a control-flow lemma.

Note that if $v_j$ and $v_k$ have different numbers of data flow dependencies or different numbers of control dependencies, these instructions cannot be shown equivalent with \approach.
We next make the notion of lemmas more precise, considering data- and control-flow dependencies in turn.

\subsubsection{Data Flow Lemmas}

Data flow lemmas are based on the operands of the instructions $v_j\in f$ and $v_k \in g$.
In general, the operands to an instruction cannot be rearranged without changing the semantics of the instruction.
For example, \cinline{strcmp(a, b)} $\ne$ \cinline{strcmp(b, a)}.
Thus, \approach builds data flow lemmas by positionally proposing that the $i$th operands are equivalent.

Some instructions have no operands (primarily function calls with no arguments).
These instructions receive a lemma of the form ``$\text{true} \implies v_j = v_k$'' which is counted, for the purposes of computing the weight, as a single data flow dependency.

\subsubsection{Control Flow Lemmas}

\approach also generates lemmas based on the control flow conditions under which instructions execute.
In general, control dependencies are defined at the basic block level, rather than the instruction level, though \approach's proof engine operates on the instruction level.
As a result, if basic block A depends on basic block B, we say that all instructions in A depend on the instruction in block B which induced the control flow (which is necessarily a branch instruction, like \cinline{loop} or \cinline{if}).
In Fig.~\ref{fig:inference}, $\text{loop}_f=\text{loop}_g \implies +_{(3),f}=+_{(5),g}$ is a control-flow lemma.
The \cinline{+} instruction on line 3 of $f$ is in a basic block that is control dependent on the block ending with the \cinline{loop} instruction in $f$. 
Likewise, the \cinline{+} instruction on line 5 of $g$ is in a basic block that is control dependent on the block ending with the \cinline{loop} instruction in $g$.
Combining these forms the lemma.
Control dependencies are \emph{labeled} with the branch decision (true or false) at each branch statement.
To ensure that \approach models the control flow behavior of each function, \approach only generates lemmas when they follow the same branch decision (e.g. are both true or both false).

A basic block may have more than one control dependency.
It is not immediately clear how to build lemmas when one or more of the instructions have multiple control dependencies.
If a pair of instructions each has $n$ control dependencies, then there are $n^2$ possible combinations.
Intuitively, it is not the case that each control dependency must be equivalent to each other control dependency.
Rather, some subset must be equivalent.

In this work, we develop a novel method of \emph{ordering} control dependencies such that two instructions can only align if their control dependencies align in that order.
If the control dependencies of instructions $p_j\in f$ and $p_k\in g$ are placed in order, then the only way for $p_j$ and $p_k$ to be equivalent is if the $i$th control dependencies are each equivalent to each other.
Ordering control dependencies significantly decreases the number of combinations of dependencies that have to be examined and allows one to know exactly which dependencies need to be compared to show to instructions equivalent.

To show the validity of our ordering strategy, we begin with several definitions.
Let A, B, C be distinct basic blocks in $f$, and A', B', and C' be distinct basic blocks in $g$.

The definition of alignment states that a basic block A in function $f$ aligns a basic block A' in function $g$ if and only if its operators are equivalent and its dependencies align.
Here, we are concerned with control dependencies; each control dependency of A must align with a control dependency of A'.
Alignment of control dependencies is a necessary condition for alignment.
Formally if $D_A$ is a control dependency of A, and $D_{A}'$ is a control dependency of A',
\begin{equation}
A\text{ aligns }A' \implies \left(\forall D_A \exists D_{A}' \text{ s.t. }D_A\text{ aligns }D_{A}'\right) \wedge \left(\forall D_{A}' \exists D_A\text{ s.t. } D_{A}\text{ aligns }D_{A}'\right)
\end{equation}

Next, we define indirect (transitive) dependence.
A is indirectly (transitively) dependent on C if A is dependent on C or if any dependency B of A is indirectly dependent on C.
In other words, A is indirectly dependent on C if there exists a path in the control dependence graph from A to C.
For brevity, we use ``dep'' to denote ``depends on'', and ``idep'' to denote ``indirectly depends on''.
Second, we make use of a depth-first node order (denoted ``$\text{dfo}: \text{BasicBlock}\rightarrow \mathbb{N}$''), also known as a reverse-post order.
This dfo is defined such that the left branch is less than the right branch.
The opposite convention can be obtained by reversing the order in which successors are visited when building the depth-first spanning tree for the dfo.
Finally, for brevity, we use the phrase ``A aligns A' '' to mean ``the instructions in A and A' '' align\footnote{If A is a dependency to another block, only the branch instructions needs to align for that block to align.}.

We define < for ordering control dependencies as:

\begin{equation}
C < B = 
\begin{cases}
      \text{dfo}(C) < \text{dfo}(B), & \text{if}\ C\text{ idep }B = B\text{ idep }C \\
      false, & \text{if}\ C\text{ idep }B \wedge \neg (B\text{ idep }C) \\
      true, & \text{if}\ B\text{ idep }C \wedge \neg (C\text{ idep }B) \\
    \end{cases}
\label{eq:controllt}
\end{equation}
We assert that, if $A\text{ dep }B, C$ and $A'\text{ dep }B', C'$:
\begin{equation}
B\text{ aligns }B' \wedge C\text{ aligns } C' \implies (C < B \wedge C' < B') \vee (B < C \wedge B' < C')
\label{eq:ordering}
\end{equation}

In other words, ordering control dependencies according to $<$ as defined in Equation~\ref{eq:controllt} is a necessary condition for aligning instructions in basic blocks A and A'. We first prove two useful facts which we will need to show Equation~\ref{eq:ordering}.

For the first fact, let X, Y and Z be basic blocks in $f$, and let W be a basic block in $g$. Further, let Z idep X, Y, and let X and Y be not necessarily distinct. Then:

\begin{equation}
X\text{ aligns }W\wedge Y\text{ aligns } W \implies X = Y
\label{eq:control_dependency_uniqueness}
\end{equation}

\begin{proof}[Proof of Equation~\ref{eq:control_dependency_uniqueness} by contradiction and induction]
Assume X aligns W, and Y aligns W, but $X \ne Y$ (that is, they are distinct).
By the definition of alignment, each control dependency of X aligns with a control dependency of W.
The same is true for Y and W.
This means X and Y have the same number of control dependencies.
Then we have:

\textbf{Base Case: X and Y have no control dependencies.}
By definition, if X and Y have no control dependencies, then control always flows through X and Y.
In other words, all paths through the CFG pass through X and Y, including all paths that pass through Z.
Either X or Y must be executed first, since they are distinct.
Say X is executed first. (The argument for Y is symmetrical).
Because Z is indirectly dependent on X, by definition, there is a sequence of control dependencies $D_1\dots D_k$ such that Z dep $D_1$, $D_i$ dep $D_{i+1}$, and $D_k$ dep X.
Y is an indirect dependency of Z, but Y cannot be in any such $D_1\dots D_k$ because Y has no control dependencies.
So Y must be on another path from X to Z.
But Y this means that Y is indirectly control dependent on X, and Y has no control dependencies.
We have reached a contradiction, so it must be the case that when X and Y have no control dependencies, X = Y.

\textbf{Inductive Step: X and Y have control dependencies.}
By inductive assumption, for all dependencies $D_X$ of X there exists a dependency $D_Y$ of Y such that $D_X = D_Y$.
We call such a shared dependency D.
Because D is a control dependency of X, X there must be a CFG path from D to X that is postdominated by X.
Y cannot be in that path, because X postdominates all bocks on that path; if Y were postdominated by X, it would not be an indirect control dependency of Z.
Likewise, there is a path postdominated by Y from D to Y that does not contain X.
This means that the paths D to X and D to Y must meet at D and are distinct.
Thus, they are entered by different branch conditions (i.e. one is the ``true'' branch from D, the other is the ``false'' branch).
But X aligns W and Y aligns W, which means that W is dependent on both the true and false branches of the block in $g$ aligned with D.
This is a contradiction, and so X = Y when X and Y have control dependencies.
This completes the inductive step.
\end{proof}

Using this, we show:

\begin{equation}
B\text{ idep } C \wedge \neg (B' \text{ idep } C') \wedge C\text{ aligns }C' \implies \neg (B\text{ aligns }B')
\label{eq:control_isomorphic}
\end{equation}
\noindent
\begin{proof}[Proof of Equation~\ref{eq:control_isomorphic} by contradiction]
Suppose $B\text{ idep } C \wedge \neg (B' \text{ idep } C) \wedge C\text{ aligns }C'\wedge B\text{ aligns }B'$.
Because $B\text{ idep } C$, by definition, there exists some sequence of dependencies $D_1,...,D_k$ between B and C such that $B\text{ dep }D_1$, $D_i\text{ dep }D_{i +1}$, and $D_k\text{ dep } C$.
Then, because B\text{ aligns }B', each dependency of B must align with a dependency of B'.
So there must exist some $D'_1$ that aligns with $D_1$.
Likewise, $D_1$ can only align with $D'_1$ if all of their dependencies align, and so on.
So there is a sequence of $D'_i$ such that $D_i$ aligns $D'_i$
Again by the definition of alignment, there must be some dependency $E'$ of $D'_k$ that aligns with $C$.
By Equation~\ref{eq:control_dependency_uniqueness}, $E'$ must be $C'$.
(In Equation~\ref{eq:control_dependency_uniqueness}'s terms, $W=C$, $X=C'$, $Y=E'$, $Z=D_k'$).
So there exists a sequence of control dependencies from B' to C', which means B' idep C'.
However, $\neg(B' \text{ idep }C')$.
We have reached a contradiction, so $\neg(B\text{ aligns }B')$.
\end{proof}

With this, we can now prove that ordering according to Equation~\ref{eq:controllt} is a necessary condition for alignment.

\begin{proof}[Proof of Equation~\ref{eq:ordering}]
We assume $B\text{ aligns }B' \wedge C\text{ aligns } C'$.
To show that the $ (C < B \wedge C' < B') \vee (B < C \wedge B' < C')$, we must use Equation~\ref{eq:controllt}, the definition of $<$.
Equation~\ref{eq:controllt}'s conditions are expressed in terms of the dependence relationships among the dependencies of A and A'.
 We consider all combinations of dependence relationships between B and C, as well as all possible dependence relationships between B' and C'.
By Equation~\ref{eq:control_isomorphic}, however, we need only consider cases where the dependence relationship between B and C matches the dependence relationship between B' and C'. 
(Likewise, the dependence relationship between C and B must match the dependence relationship between C' and B').
Otherwise, by Equation~\ref{eq:control_isomorphic}, we have $\neg(B\text{ aligns }B')$ or $\neg (C\text{ aligns }C')$, which is a contradiction.

\noindent
\textbf{Case 1: $\neg(C\text{ idep }B) \wedge B\text{ idep }C$ and $\neg (C'\text{ idep } B')\wedge B'\text{ idep }C'$}

Plugging this set of conditions into Equation~\ref{eq:controllt} means choosing the third case, which means C < B.
The same is true for B' and C'.
So $C < B\wedge C' < B'$.

\noindent
\textbf{Case 2: $C\text{ idep }B \wedge \neg(B\text{ idep }C)$ and $C'\text{ idep } B' \wedge \neg(B'\text{ idep }C')$}

Plugging this set of conditions into Equation~\ref{eq:controllt} means choosing the second case, which means $B < C \wedge B' < C'$.

\noindent
\textbf{Case 3: $\neg(C\text{ idep } B) \wedge \neg(B\text{ idep }C)\wedge \neg (C'\text{ idep }B') \wedge \neg(B'\text{ idep }C')$}

First, we show that B and C have a most recent common indirect control dependency D.
There exists a path from the entry block to A that goes through B.
Likewise, there exists a path from the entry block to A that goes through C.
These paths must both contain the entry block, so such a common ancestor in the CFG always exists.
Because B is a control dependency of A, there exists a path from B to A postdominated by A (except B).
C cannot be on this path, because C, as a control dependency, is not postdominated by A. Likewise, B cannot be on the path from C to A.
Therefore, the paths from the common control flow ancestor through B and C to A differ by at least one node.
This means at some node, the paths diverge.
This node must be an indirect dependency of B and C, because it controls whether control can flow to B or C.

The above also true for A', B', and C'.
We denote the common direct descendant of A' and B', D'.

B aligns B', so the dependencies of B must align the dependencies of B'.
By Equation~\ref{eq:control_dependency_uniqueness}, all such dependencies are unique.
Further, this must be true of the dependencies of the dependencies of B, and their dependencies, recursively.
This includes D, which means D align D'.
Because D aligns D', either B and B' are both reached via the true branch of D and D', or the false branch.
C and C' must then each be reached following the other branch condition.

Thus, either $\text{dfo}(B) < \text{dfo}(C) \wedge \text{dfo}(B') < \text{dfo}(C')$ or $\text{dfo}(C) < \text{dfo}(B) \wedge \text{dfo}(C') < \text{dfo}(B')$.
In turn, by Equation~\ref{eq:controllt}, $B < C \wedge B' < C'$ or $C < B \wedge C' < B'$.

\noindent
\textbf{Case 4: $C\text{ idep } B \wedge B\text{ idep }C\wedge C'\text{ idep }B' \wedge B'\text{ B' idep C'}$}

In \approach's loop-proof system, dependencies corresponding to forward edges must be aligned before dependencies corresponding to back-edges.
Further, the dfo for the forward edge is necessarily less than the one corresponding to the back edge.
Either C must have been aligned with C' first or B must have been aligned with B' first.
If C was aligned first, we have $\text{dfo}(C) < \text{dfo}(B)\wedge\text{dfo}(C')<\text{dfo}(B')$, so by Equation~\ref{eq:controllt}, $C<B\wedge C' < B'$.
Conversely, if B was aligned with B' first, then we have $\text{dfo}(B) < \text{dfo}(C)\wedge\text{dfo}(B')<\text{dfo}(C')$, so by Equation~\ref{eq:controllt}, $B<C\wedge B' < C'$.

\end{proof}

\subsubsection{Function Pointers} The preceding discussion assumes a distinction between values and instructions, but this is not always the case when functions can be passed as values (such as with function pointers). Calls to function pointers don't have a statically defined operator.
We consider function pointers to have equivalent operators if the values they call have themselves been proven equivalent; otherwise, they are treated like other instructions.

\subsection{Loops}
\label{sec:loops}

\begin{wrapfigure}{R}{5cm}
\centering
\includegraphics[width=100pt]{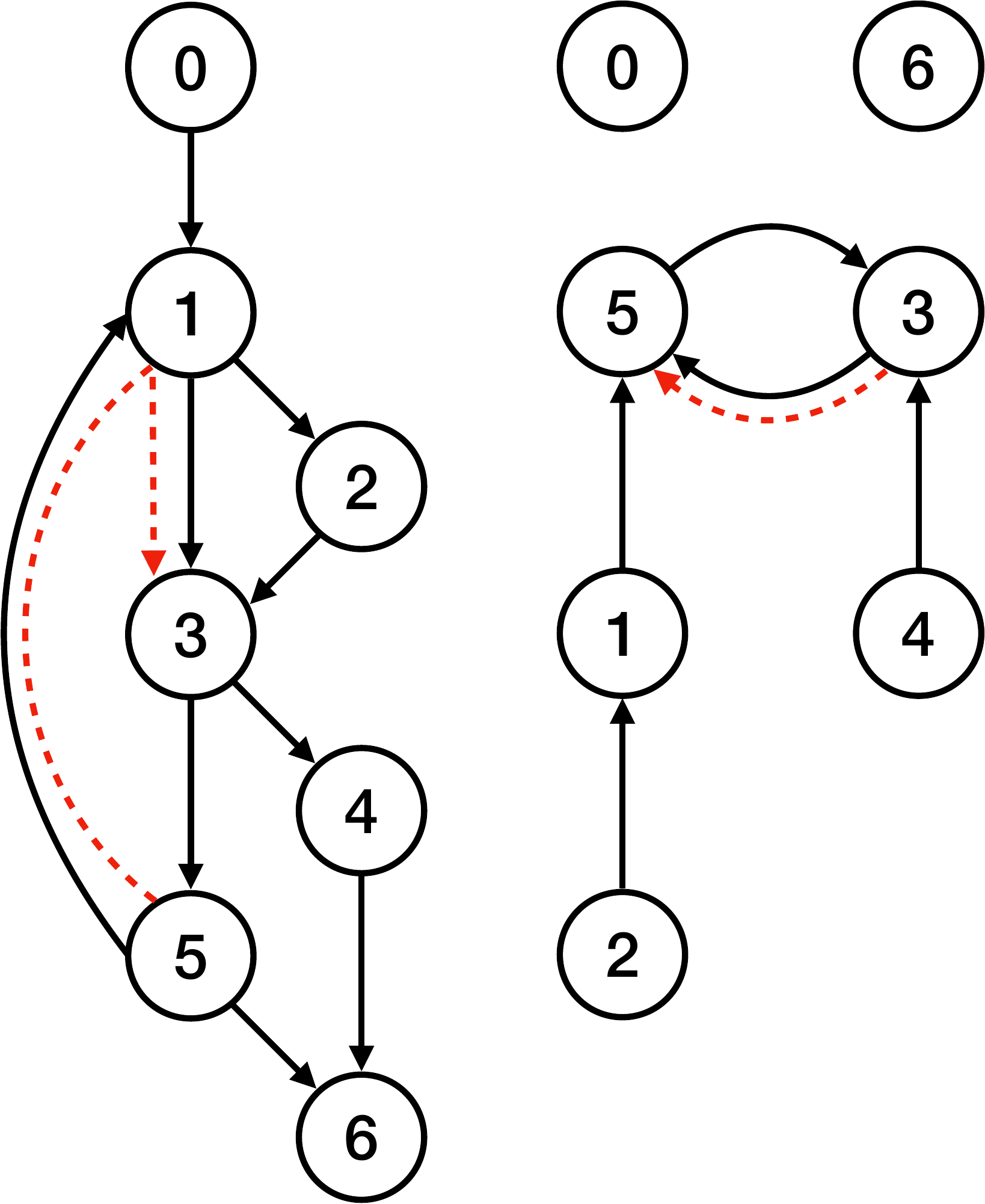}
\caption{\small A CFG (left) and CDG (right). There is a cycle of control dependencies between blocks 5 and 3. The red dashed line indicates a back-dependency, as does the CDG edge from 1 to 5.}%
\label{fig:cyclic_dependency}
\end{wrapfigure}

We calls cycles in the control flow graph loops.
Like many program analysis tools, \approach assumes reducible flow graphs, which means all loops must be \emph{natural loops}.
Natural loops have a single basic block at which the loop is entered, called the head.
All natural loops have a \emph{back-edge}, which is an edge from B to A such that A comes before B in a topological sort of the control flow graph.
In Fig.~\ref{fig:cyclic_dependency}, the CFG edge from blocks 5 to 1 is a back edge.

Loops may result in cycles in of dependencies.
These cycles may contain only data flow dependencies, only control flow dependencies, or both control and data flow dependencies.
For instance, when mutable variables are updated in the body of a loop, a cycle of dataflow dependencies is formed.
In Fig.~\ref{fig:inference_a}, this occurs in both functions when the $i$ variable is incremented.
A loop with a break statement inside an if statement may result in a cycle of control dependencies: the loop branch is control dependent on the if branch, and the if branch is control dependent on the loop branch.
Fig.~\ref{fig:cyclic_dependency} shows the CFG for such a loop.
And in \cinline{do...while} loops, the head of the natural loop is control dependent on the block with the loop's branch statement, at the end of the loop.
At the same time, the branch statement may be computed using values in the head of the natural loop.
This creates a cycle of both control and dataflow dependencies.

Cycles of dependencies result in cycles of lemmas, which make it impossible to prove two loops equivalent unless special consideration is given to loops.
In Fig.~\ref{fig:inference} to prove the \cinline{+} instructions from the desugared \cinline{i++} expressions equivalent, the $\phi$ instructions must be shown equivalent.
But the values of those same \cinline{+} instructions are operands to the $\phi$ instructions.

To resolve this, \approach identifies \emph{back-dependencies}: dependencies that can only be reached from the dependent block by following the loop's back-edge when traversing the CFG.
Fig.~\ref{fig:cyclic_dependency} highlights such a dependency.
(Intuitively, this can be thought of as dependency edges where the dependency is "above" the dependent block in the code).
Data flow back-dependencies necessarily end in a $\phi$ instruction at the loop's head.
\approach examines each operand of each $\phi$ instruction at the head of a loop to determine if it is a back-dependency.
Unlike data flow dependencies, control back-dependencies do not necessarily end at the loop's head.
The highlighted dependency in Fig.~\ref{fig:cyclic_dependency} is an example.

\approach uses induction to bootstrap proofs of loops.
As discussed in Section~\ref{sec:definitions}, the base case(s) are the dependencies from outside the loop.
For the inductive step, \approach assumes that the $i$th loop iterations are equivalent, and attempts to prove the $i + 1$st iterations equivalent.
The necessary assumptions are precisely the back-dependencies previously identified.
In Fig.~\ref{fig:inference}, there is a cycle of dependencies between the $\phi$ instructions and the $+_{(2),f}$/$+_{(6),g}$ instructions.

Like any other dependency, each back-dependency is modeled as a lemma in \approach's logical system.
For each back-dependency lemma of the form $A \implies B$ with weight $1/n$, \approach adds an Equation~\ref{eq:smalllemma}-form lemma of the form $\text{true} \implies B$ with weight $1/n$.
For Fig.~\ref{fig:inference}, that's  $+_{(2),f}=+_{(6),g} \implies \phi_f=\phi_g, 1/2$ and $\text{true} \implies \phi_f=\phi_g, 1/2$, respectively.
This means the total provable weight of proposition $B$ is greater than $1$, and that $B$ can be proven without using any back-dependency lemmas.
Adding these lemmas is the way \approach assumes the inductive hypothesis.
In Fig.~\ref{fig:inference}, the base case is completed by satisfying the condition of $0=0 \implies \phi_f=\phi_g, 1/2$, while inductive hypothesis is completed by satisfying $\text{true} \implies \phi_f=\phi_g, 1/2$.
Then, to complete the inductive step, \approach attempts to satisfy the conditions of the back-dependency lemmas.
If it can, then it has successfully inductively proven the loop; otherwise, it is unable to show the loops equivalent.
In this case it can; using the facts that $\phi_f=\phi_g$ and $1=1$, $+_{(2),f}=+_{(6),g}$.
This satisfies the condition of the back dependency lemma $+_{(2),f}=+_{(6),g} \implies \phi_f = \phi_g, 1/2$.
If \approach is unable to show the loops equivalent, it recursively revokes lemmas and propositions proven based on the assumptions.
The mechanism by which this happens is explained in more detail in Section~\ref{sec:inference}.

\begin{wrapfigure}{R}{0.50\textwidth}
\begin{minipage}{0.50\textwidth}
\begin{algorithm}[H]
\SetKwInOut{Input}{input}
\SetKwInOut{Output}{output}
\KwData{lemmas: dict[Equivalence, list[Equivalence]]}
\KwData{baseCases: list[Equivalence]}
\KwData{loopAssumtions: dict[Equivalence, list[Equivalence]}

worklist $\gets$ Queue(baseCases)

proofGraph $\gets$ ProofGraph(baseCases)

\While{worklist is not empty}{
  condition $\gets$ worklist.pop()
  
  conclusions $\gets$ lemmas[condition]
  
 \For{conclusion \textbf{in} conclusions}{
   weight $\gets$ proofGraph.addEdge( condition, conclusion)
   
   \If{weight = 1.0}{
       worklist.put(conclusion)
   }
 }
}
 \For{condition \textbf{in} loopAssumptions.keys()}{
   conclusions $\gets$ loopAssumptions[condition]
   
   \For{conclusion \textbf{in} conclusions}{
      proofGraph.removeEdge( condition, conclusion)
   }
 }
\Return proofGraph.validNodes()
\vspace{4pt}
\caption{\small \approach's inference algorithm. An Equivalence object represents a proposition that two values are equivalent. Lemmas are represented as a dictionary, with the lemma condition as the key and the conclusion the value. We group lemmas with the same conclusions in a list.}
\label{alg:inference}
\end{algorithm}
\end{minipage}
\vspace{-0.5cm}
\end{wrapfigure}

\subsection{Inference}
\label{sec:inference}

After all lemmas have been generated, \approach begins proving values equivalent by induction (at the instruction level, rather than the loop level as described in Section~\ref{sec:loops}).
The base cases consist of facts which are known or assumed to be equivalent a priori.
Function parameters in the same positions are assumed to be equivalent, even if they have different names.\footnote{We ignore types when matching function arguments. See Section~\ref{sec:limitations}.}
In doing so, \approach ignores the code authors' \emph{intentions} of what should be passed to each parameter and models what \emph{would} happen if the same parameters are passed.
Similarly, global variables are assumed to be equivalent if they have the same name.
Identical constant values are considered equivalent.

Lemmas are represented in the form shown in Equation~\ref{eq:smalllemma}.
This allows them to be easily indexed in a hash table by their condition.
A pair of values is proven equivalent when that proposition has total accumulated weight 1.0.
When this happens, finding the lemmas whose conditions are subsequently satisfied is a simple dictionary lookup.

\approach builds its proof using a worklist algorithm as shown in Algorithm~\ref{alg:inference}.
The base cases are inserted into a queue, the worklist.
Then, at each iteration, a proposition is popped from the worklist.
If the equivalence satisfies the condition of any lemmas, the conclusions of those lemmas are added to the worklist.
The process repeats until the worklist is empty.
Whenever a proposition a lemma's conditions are satisfied, it is added to a proof-graph data structure.
The nodes in the proof graph are propositions---pairs of equivalent values---and the edges are lemmas that relate one proposition to another.
When the worklist is empty, \approach revokes any lemmas added to enable the induction of loops and decreases the weight associated with each corresponding conclusion.
Then, if the back-dependency lemmas were not proven, the weight of the nodes assumed true will drop below 1.0.
When this happens, \approach removes each lemma and proposition from the proof-graph that is reachable from the assumed node, thus removing any propositions proved based on the failed loop induction.
Any remaining proposition that is still true in the graph becomes a part of the alignment.
An example proof graph after attempting to align $f$ and $g$ is shown in Fig.~\ref{fig:inference_d}.

\subsection{Options}
\label{sec:options}

\approach has several settings that can modify its behavior.
We find it useful to allow for partial proofs of loops.
Even if the beginning of two loops are equivalent, any differences within the loop will result in the whole loop failing to align.
As a result, \approach also offers an option called ``partial loop'' mode.
In this mode, inductive loop assumption lemmas are generated, but back-dependency loop edges are not.
Further, loop-assumption lemmas are not revoked (in Algorithm~\ref{alg:inference}, the set of nested for loops before the return are ignored).

Additionally, control dependencies can be disabled, in which case \approach uses only data flow dependencies to perform its alignment.
This makes \approach more flexible in what it can align.
We don't disable control dependencies in any experiment in the paper.

Using either of these options makes \approach unsound, but can still be useful. 

\section{Evaluation}
\label{sec:evaluation}

In this section, we characterize the capabilities of \approach.
First, in Section~\ref{sec:correctness} we substantiate our claim that \approach's dependency-based equivalence is sound, and describe the limitations of this claim when applied to real code.
Next, we show how \approach performs on full-featured C, illustrating the applicability of dependency-based equivalence.
While there do exist tools that solve related problems~\cite{DSE,symdiff,ardiff}, we are unaware of any existing tool which produces equivalence alignments against which we can directly compare \approach.
Instead, we appeal to symbolic execution, a widely-used technique in program analysis and equivalence work.
In Section~\ref{sec:symexe}, we describe how we use symbolic execution to build approximate reference alignments and compare against them.
While these are both unsound and incomplete, they serve as a useful reference to more intuitively characterize \approach's behavior.
Finally, in Section~\ref{sec:runtime}, we measure \approach's runtime performance.

In the next section, we use \approach to solve a real-world task: checking the correctness of predictions produced by a machine learning model and using the alignment to enable an evaluation of variable names.

\subsection{Correctness}
\label{sec:correctness}

\approach is sound, subject to some well-defined limitations.
We claim that instructions that \approach aligns have functionally equivalent values, as defined in Section~\ref{sec:definitions}; however, \approach is necessarily incomplete and will not find all functionally equivalent values.

\subsubsection{Soudness}
\label{sec:soundness}
By construction, the set of aligned nodes \approach produces are isomorphic in terms of their dataflow graphs and their control dependence graphs.
This is because in order to align, all dependencies of a pair of instructions must align, so if an node is added to the alignment, all of the edges to it from nodes already in the graph are added along with it.
Programs with isomorphic program dependence graphs and SSA forms are semantically equivalent~\cite{yang1989detecting}.
Therefore, \approach is sound.

Computing graph isomorphisms is a difficult problem, and there are no known polynomial time algorithms.
It is not known if finding a graph isomorphism is NP-complete~\cite{graph_isomorphism_problem}.
However, \approach is able to do this efficiently by taking advantage of the characteristics of SSA form and control dependence graphs to significantly reduce the search space of combinations.\footnote{There exist efficient algorithms for determining if planar graphs are isomorphic, but data flow graphs are not necessarily planar. It is easy to write code with a $K_{3,3}$ data flow sub-graph.}

\subsubsection{Limitations}
\label{sec:limitations}
\cite{yang1989detecting}'s theorem is defined only for programs meeting certain characteristics; \approach inherits these limitations.

\textbf{Side Effect Ordering:}
\approach considers the equivalence of instruction side effects independently from the side effects of other instructions.
For instance, if given \cinline{printf("A"); printf("B");} with \cinline{printf("B"); printf("A");}, \approach would align the \cinline{printf("A")}s and \cinline{printf("B")}s.
While each of these instructions are indeed equivalent individually, the two programs write different output to the screen.
\approach could be extended to differentiate instructions based on their side effects by adding edges between instructions that produce side effects if there exists a path on which both could be executed.
\approach assumes that program state at the start of executing each function is the same.
(Otherwise, even textually identical functions could produce different results.)
What makes \approach unsound with respect to side effects is the possibility of executing instructions with side effects in different orders, changing some global state in a different way or producing different output.
Side-effect dependencies would eliminate this risk, though some of the power of \approach is in its ability to flexibly align instructions out of the order they appear in the code text.
In principle, any function call could have arbitrary side effects, drastically limiting flexibility.
Modifications to mutable data structures are modeled as function calls with side effects in \approach; these at least would only need to have dependencies between operations on the same data structure.
Finally, if the side effects are reflected in control or data flow or the instructions otherwise have different dependencies, as is often the case, \approach will remain sound.

\textbf{Types:}
\approach does not presently consider types when building an alignment.
In practice, different types are usually operated on by different instructions, so this does not have significant practical effect.
In the worst case, overloaded operators (like \cinline{+}) can in principle lead to nonsensical lemmas containing propositions asserting values of two different types are equivalent.
\approach could be extended to build on the frontend language's existing type inference system to infer types and then use them to rule out nonsensical lemma construction.

Finally, like many other program analysis techniques, \approach requires a reducible flow graph because it relies on natural loop analysis.
However, even in C, one of the few programming languages that can form irreducible CFGs, they are extremely rare~\cite{stanier2012study}.

\subsection{Comparison with a Symbolic-Execution-Based Alignment}
\label{sec:symexe}

We use symbolic execution to build reference alignments against which \approach can be compared.
Symbolic execution provides a reference against a widely-known technique to characterize codealign's behavior, showing the validity of dependency-based equivalence in practice.

The idea behind symbolic execution is to exhaustively test a program by considering all inputs simultaneously.
To do this, program inputs are defined as symbolic variables instead of specific (concrete) values.
The intermediate values produced as the program executes are represented as mathematical expressions defined in terms of the symbolic input variables.
When a symbolic variable is used to decide a branch, symbolic execution may explore both paths.

To build our symbolic-execution alignments, we execute each program we are comparing symbolically, logging the symbolic expressions that represent the values produced each time that instruction is executed (along any path).
Using z3~\cite{z3}, an SMT solver, we then check the equivalence between each pair of instructions between the two pieces of code.
Because these symbolic expressions represent the computation of these values on all possible inputs, if the symbolic representations of the execution result are equivalent, the instructions are aligned (as defined in Equation~\ref{eq:functionalequivalence}).

In symbolic execution, the same instruction may be executed multiple times along different paths, including loop iterations.
There may be different values associated with each instruction across different paths, so we also define equivalence for this scenario.
In general, there will be $\ell$ executions of one instruction and $k$ executions of another.
Intuitively, each execution of one instruction should have a symbolic value equivalent to an execution of the other.
As a practical matter, we terminate symbolic execution after collecting 10,000 instruction executions.
If one loop is larger than another, it is possible that the instruction in the smaller loop is executed more times; that is, $\ell \ne k$.
To handle this issue, we require that each execution of the instruction executed fewer times must be equivalent to an execution of the other instruction to consider two instructions equivalent.
In practice, limits on execution and the need to aggregate values from multiple paths results in unsoundness and incompleteness.

\subsubsection{Experiment Methodology}

We modify the Klee symbolic execution engine~\cite{klee} to log the symbolic values of variables after the execution of instructions. 
Klee is built on top of LLVM, and functions as an LLVM IR interpreter that can represent values symbolically.

We only log symbolic values for the LLVM IR instructions that we can deterministically map to \approach instructions, and those that do not return memory addresses (e.g. \cinline{getelementptr}). Klee uses concrete memory addresses; even identical programs may have different concrete memory addresses on different runs.

We perform experiments on the POJ-104 dataset, a component of CodeXGlue~\cite{codexglue}.
The POJ-104 dataset consists of responses to programming competition questions.
It is often used as a type-4 code clone detection benchmark because responses to the same question can, in some senses, be considered equivalent.
We choose a clone detection dataset because it is somewhat more likely that responses to the same programming question contain equivalent values that it is that two randomly-selected open-source functions will; that is, the results are more likely to be interesting.
However, in this experiment, we are not trying to determine which examples are attempted solutions to the same programming question.
We are illustrating how \approach detects equivalent values inside pairs of functions.

As with many programs from programming challenges, input is read through standard input (typically through the \cinline{scanf} or \cinline{gets} functions).
In the interest of simplifying symbolic constraints, we rename each \cinline{main} function, and pass to that function the symbolic input that would have been read from standard input.
Then, we remove the call, the associated variables, and associated initialization code from the renamed main function.

Before sampling examples from the dataset, we perform several filtering steps to remove functions that Klee cannot handle or which we cannot preprocess into a form that Klee can handle without substantial alteration of the example. These include functions that use unhandled input/output functions (like C++ input methods \cinline{cin} and \cinline{cout}, or \cinline{scanf} arguments we cannot resolve to a variable); contain floating point variables that Klee automatically sets to 0.0, lack a traditional \cinline{main} method, or do not parse.
After filtering, 17,106 examples remain of the original 52,000.

\looseness-1
We sample 1100 different pairs of functions from the filtered dataset from three categories: (1) 500 pairs of different solutions to the same problem, (2)  500 pairs of identical functions (the same solution to the same problem), and (3)
100 pairs of solutions to different problems, which are expected to have very few, if any, equivalent instructions.
Although the 500 pairs of textually identical functions are trivially equivalent, both \approach and symbolic execution ignore textual similarity.

We then compile each example to LLVM IR and symbolically execute them, logging the symbolic value produced by executing each instruction in the rewritten \cinline{main} functions, until they log 10,000 constraints or reach a timeout of one hour.
We then build the alignment by comparing the constraints for each instruction in one function with each instruction in the other, with a timeout of 2 hours.
Finally, we map LLVM IR instructions to \approach instructions and compare the alignments.

\subsubsection{Results}

\begin{wraptable}{R}{0.36\textwidth}
\centering
\begin{tabular}{lr}
\toprule
& Precision \\
\midrule
Self-Alignment & 99.9\% \\
Same Problem & 95.9\% \\
Different Problem & 100.0\% \\
\bottomrule
\end{tabular}
\caption{\small agreement between \approach and symbolic execution.}
    \label{tab:symexe}
\end{wraptable}

Of the experiments, 737/1100 succeed.
The errors, by cause, are:
(1) 144 have constraints for fewer than five instructions,
(2) 69 have more than 200 MB of symbolic constraints,
(3) 46 contain floating point instructions despite not containing any \cinline{float} variables,
(4) 46 contained symbolic variables that are not a part of the functions' parameters (so we could not map them between functions in the pair),
(5) 30 failed to compile,
(6) 23 timed out after 2 hours,
(7) 4 had issues mapping LLVM IR instructions to \approach instructions,
(8) \approach crashed on 1 example.

The results are shown in Table~\ref{tab:symexe}.
The majority of pairs of instructions between the functions are, as one may expect, nonequivalent.
Therefore, we report precision scores.
Because we are comparing \approach against a well-understood approach, we define the symbolic execution alignment as the ground truth for the purposes of calculating precision, though it is unsound and incomplete.

We evaluate over all possible combinations of values.
Precision scores are very high meaning that instruction pairs \approach determined were equivalent had equivalent symbolic values.
There are very few exceptions: only 12 of the 737 functions contained any false positives.
We manually analyzed all twelve cases and identified three reasons for the differences.
The most common reason, affecting 6/12 of the cases, were the use of different integer types (e.g. \cinline{int} vs \cinline{long}).
While these types did indeed hold equivalent values, Klee constraints are modeled with bitvectors, and so different integer types are automatically nonequivalent.
The second most common reason, affecting 5/12 cases, is when a mutable data structure is mutated and accessed twice; \approach considers the accesses the same when they are actually different (See Section~\ref{sec:limitations}).
The final case has to do with uninitialized memory.
\approach treats uninitialized memory as a constant and allows it to align with other uninitialized memory; otherwise, a function with uninitialized memory could not align with itself.
In the final false-positive case, \approach aligned two different segments of uninitialized memory.

We do not report recall scores because our symbolic-execution-based alignment performs an overwhelming number of spurious alignments.
The culprit is primarily concrete values.
In symbolic execution, only the values passed to the function and other values computed based on the input are symbolic; local variables that do not interact with symbolic variables retain concrete values.
It is easy for these concrete values to spuriously correlate with other concrete values.
For instance, \cinline{printf} returns the number of characters that were printed, which may happen to be equal to an unrelated value, such as a loop's iteration counter.
The necessarily loose definitions for symbolic multi-execution equivalence defined in Section~\ref{sec:symexe} allows these to be marked as equivalent.

\subsection{Runtime Performance}
\label{sec:runtime}

\begin{figure}
    \includegraphics[width=3in]{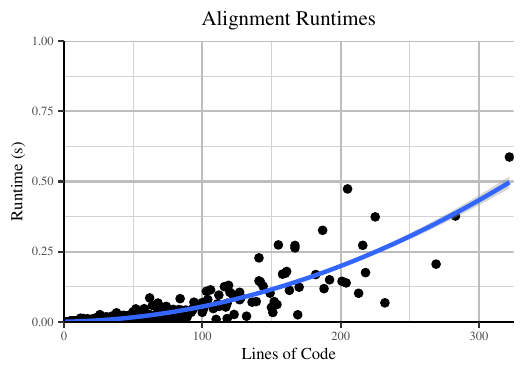}
    \caption{\small Runtime performance of \approach on functions from GNU Coreutils with a quadratic regression trend line.}
    \label{fig:runtime}
\end{figure}

\approach is very fast.
We quantify \approach's runtime performance with a benchmarking experiment on C functions in GNU Coreutils version 9.5.
We extract all functions from all files the src subdirectory, excluding functions which contain features \approach does not currently support, such as \cinline{#ifndef} macros and \cinline{goto} statements.
In total, we collected 1176 functions.
We then align each function with itself and measure the time it takes to build each equivalence alignment.
Aligning a function with itself is the worst case in terms of runtime. (When aligning two completely different functions, there are few lemmas and little induction, leading to early termination.)
We used an 18-core Intel Xeon Gold 6240 CPU with 256GB RAM, though the experiment code was single-threaded and processed each function serially.

Fig.~\ref{fig:runtime} displays \approach's runtime performance.
Despite being written in pure python, \approach can build an alignment in under a second, even on large functions spanning hundreds of lines.
This dramatically outperforms, e.g., \cite{gupta2018effective}'s technique for equivalence-checking compiler optimizations, though given the different hardware and task the comparison is not apples-to-apples.

\section{Utility: Evaluating a Neural Decompiler}
\label{sec:neuraldecompilation}

\begin{figure*}
    \begin{subfigure}[t]{0.49\textwidth}
    \begin{lstlisting}[style=cstyle,basicstyle=\ttfamily\bfseries\tiny]
int write_response(int fd, char *response, int len) {
  int retval, byteswritten = 0;
  while (byteswritten < len) {
    retval = write(fd, response + byteswritten, len - byteswritten);
    if (retval <= 0) {
      return 0;
    }
    byteswritten += retval;
  }
  return 1;
}
    \end{lstlisting}
    \caption{\small \textbf{Original} ($f$):  writes a message to a file descriptor, retrying with the remainder of the message if the call to \cinline{write} fails to write all of the message.\label{fig:original}}
    \end{subfigure}
    \hfill
    \begin{subfigure}[t]{0.49\textwidth}
    \begin{lstlisting}[style=cstyle,basicstyle=\ttfamily\bfseries\tiny]
signed long long
write_response(int a1, long long a2, int a3) {
  int v4, v5, i;
  v4 = a3;
  for (i = 0; i < v4; i += v5) {
    v5 = write(a1, (const void *)(a2 + i), 
                       v4 - i);
    if (v5 <= 0)
      return 0LL;
  }
  return 1LL;
}
    \end{lstlisting}
    \caption{\small \textbf{Decompiled}: Fig.~\ref{fig:original}, compiled and determinstically decompiled. This code is much harder to read: variable names are generic (like \cinline{a2}), variable types are inaccurate (\cinline{char *} is replaced with \cinline{long long}), and extra variables are introduced (\cinline{v4}). \label{fig:decompiled}}
    \end{subfigure}
    
    \vspace{8pt}
    
    \begin{subfigure}[t]{0.49\textwidth}
    \begin{lstlisting}[style=cstyle,basicstyle=\ttfamily\bfseries\tiny]
int write_response(int fd, char *buf, int len) {
    int i;
    for (i = 0; i < len; i += len) {
        if ((i = write(fd, buf + i, len - i)) <= 0)
            return 0;
    }
    return 1;
}
    \end{lstlisting}
    \vspace{8pt} %
    \caption{\small \textbf{Prediction} ($g$): A fine-tuned CodeT5's prediction when provided Fig.~\ref{fig:decompiled} as input. The prediction is subtlety incorrect.\label{fig:predicted}}
    \end{subfigure}
    \hfill
    \begin{subfigure}[t]{0.49\textwidth}
    
    \begin{minipage}[t]{0.59\textwidth}
    \centering
    \[\arraycolsep=1.0pt
    \begin{array}{rcl}
    \text{\cinline{<}}_{(4),f} & = & \text{\cinline{<}}_{(3),g} \\
    \text{\cinline{+}}_{(5),f} & = & \text{\cinline{+}}_{(4),g} \\
    \text{\cinline{-}}_{(6),f} & = & \text{\cinline{-}}_{(4),g} \\
    \text{\cinline{write}}_{(5),f} & = & \text{\cinline{write}}_{(4),g} \\
    \text{\cinline{<=}}_{(7),f} & = & \text{\cinline{<=}}_{(4),g} \\
    \end{array}
      \]
    \textcolor{red}{Unaligned:
    $\text{\cinline{+}}_{(8),f}$ and $\text{\cinline{+}}_{(3),g}$}
    \end{minipage}
    \hfill
    \begin{minipage}[t]{0.39\textwidth}
    \[\arraycolsep=1.0pt
    \begin{array}{rcl}
    \text{\cinline{fd}}_{f} & = & \text{\cinline{fd}}_{g} \\
    \text{\cinline{response}}_{f} & = & \text{\cinline{buf}}_{g} \\
    \text{\cinline{len}}_{f} & = & \text{\cinline{len}}_{g} \\
    \text{\cinline{retval}}_{f} & = & \text{\cinline{i}}_{g} \\
    \end{array}
    \] 
    \end{minipage}
    
    \caption{\small \emph{Left}: The partial-loop (Section~\ref{sec:options}) alignment produced between Fig.~\ref{fig:original} ($f$) and Fig.~\ref{fig:predicted} ($g$), excluding control-flow instructions. \emph{Right}: The variable name mapping derived from the alignment.}
    \label{fig:ndalignmentexample}
    \end{subfigure}
    
    \caption{\small A function (Fig.~\ref{fig:original}) from the test set in the experiment.
    Fig.~\ref{fig:decompiled} shows the same function having been compiled, then decompiled.
    Machine learning models can be used to render decompiler output more readable, but they may produce semantically nonequivalent code. 
    Fig.~\ref{fig:predicted} shows an example. 
    The alignment generated by \approach in Fig.~\ref{fig:ndalignmentexample} can be used to both detect the hallucination,  and evaluate the quality of the variable names in the decompiled function. Fig.~\ref{fig:ndalignmentexample}, shoes the use of partial-loop~(Section \ref{sec:options}) alignment to analyze why the loop does not align.
    }
    \label{fig:decompilationexample}
    \end{figure*}

Neural decompilation---generating source code from an executable binary or a deterministically derived representation of one---has become an area with rapidly increasing interest~\cite{Katz2018,coda,wu2022dnd,cao2022boosting,liang2021neutron,hudegpt,nova,slade,llm4decompile}.
Deterministic decompilers fail to recover many of the abstractions, like names and types, that make code readable (Fig.~\ref{fig:original} vs~Fig.~\ref{fig:decompiled}), because those abstractions are discarded during compilation.
Neural decompilers can help rewrite code to be more readable, but can also hallucinate, as Fig.~\ref{fig:predicted} shows.

In this section, we show how to use \approach to quantify the correctness of model predictions.
In neural decompilation, the ideal decompilation is identical to the original source.
Further, we use the alignment generated by \approach to map variables in the model's prediction with the ground truth in the original code.
This allows us to evaluate the quality of the generated variable names.
Generating these abstractions is a key reason to use neural decompilation in the first place; evaluating their quality is thus an important part of any neural decompilation evaluation.

\subsection{Methodology}

To demonstrate how \approach can be used to evaluate neural decompilers, we design a controlled experiment between two different neural decompilers.
Because language models are increasingly popular for neural decompilation (in addition to many other tasks), we fine-tune two language models, CodeT5~\cite{codet5} and CodeT5+~\cite{codet5+}, on the same training data to perform a neural decompilation task.

As a prerequisite to fine-tuning the models, we built a dataset of input/output pairs. The input to the model should be a representation of the code that can be recovered entirely deterministically from an executable form of the code. The output is the original source code written by developers. We generate this data by cloning and compiling C repositories from GitHub using the GitHub Cloner-Compiler tool\footnote{https://github.com/huzecong/ghcc} (GHCC). We then decompile the programs using Hex-Rays,\footnote{https://hex-rays.com/decompiler/} an industry-standard deterministic decompiler.
We then match the decompiled function and original function by name (using debug symbols), since in C, names uniquely identify functions.

We collect a sample of 100,000 decompiled/original function pairs, then split them into training, validation, and test sets using an $80:10:10$ split. We fine-tune CodeT5~\cite{codet5} and CodeT5+~\cite{codet5+} (220M parameter version) on this data. Finally, we generate predictions for both models on each test set example, and evaluate those predictions with \approach.

\subsection{Results}
\label{sec:neuraldecompilation:results}

The results are shown in Table~\ref{tab:neuraldecompilationresults}.
Table~\ref{tab:ndalignment} shows a \approach-based evaluation of the two models.
We say a prediction is \emph{perfectly aligned} with the reference if every value in the prediction is aligned with one in the reference, and vice versa.
Being perfectly aligned is a strong indicator of correctness.
(See Section~\ref{sec:limitations} for limitations.)
\approach shows that both models' predictions are perfectly aligned a small minority of the time (8.56\% and 13.1\% for CodeT5 and CodeT5+, respectively).
In fact, these models produce syntactically or semantically invalid code more often than they produce correct results.
Fig.~\ref{fig:predicted} shows CodeT5's incorrect prediction based on Fig.~\ref{fig:decompiled}.
If the call to \cinline{write} succeeds in writing the whole message at once, then the prediction works in the same way as the original in Fig~\ref{fig:original}.
However, if the call to \cinline{write} writes fewer bytes than expected, the prediction will silently fail to write the rest of the message while returning \cinline{1} to indicate success.
Accordingly, Fig.~\ref{fig:original} and Fig.~\ref{fig:predicted} do not align.
When using partial-loop mode (Section~\ref{sec:options}), the alignment shown in Fig.~\ref{fig:ndalignmentexample} reflects the fact that the instructions executed on the first pass through loop are correct, but that the \cinline{+}  instruction that calculates the offset in the buffer for the retry is not correct.

\begin{table*}
    \begin{subtable}{1.0\textwidth}
    \centering
    \caption{\small Alignment Results}
    \label{tab:ndalignment}
    \begin{tabular}{lrrrr}
    \toprule
     & \multicolumn{2}{c}{\emph{Functional Correctness}} & \multicolumn{2}{c}{\emph{Variable Name Quality}} \\
             Base Model & Perfectly Aligned & Invalid Code & Accuracy & VarCLR\\
    \midrule
              CodeT5 & 8.56\% & 23.6\% & 27.9\% & 0.668 \\
              CodeT5+ & 13.1\% & 21.4\% & 37.3\% & 0.717 \\
    \bottomrule
    \end{tabular}
    \end{subtable}
    
    \vspace{4pt}
    \begin{subtable}{1.0\textwidth}
    \centering
    \caption{\small Existing similarity metrics}
    \label{tab:similarity}
    \begin{tabular}{lrrrr}
    \toprule
    Base Model& CodeBLEU& CodeBERTScore & CrystalBLEU & Corpus BLEU \\ 
        & {\small \cite{codebleu}}  & {\small \cite{codebertscore}} & {\small \cite{crystalbleu}} & {\small \cite{bleu}} \\
    \midrule
    CodeT5 & 0.510 & 0.868 & 0.249 & 0.323 \\
    CodeT5+ & 0.597 & 0.890 & 0.340 & 0.424 \\
    \bottomrule
    \end{tabular}
    \end{subtable}
    
    \caption{\small \approach and several code similarity metrics used to evaluate two neural decompilers. Results exclude \approach failures, which occur in 2-3\% of cases. VarCLR was introduced by~\cite{varclr}.}
    \label{tab:neuraldecompilationresults}
    \end{table*}

Alignments generated by \approach can be further used to evaluate the quality of variable names produced.
If two aligned instructions assign their results to variables in the code, we record those pairs of variables.
With the variables aligned, we can score them using various metrics.
Here, we use exact-match accuracy and the variable name similarity metric VarCLR~\cite{varclr}.
The scores in Table~\ref{tab:ndalignment} also include parameter names, matched positionally.
The variable alignment for Fig.~\ref{fig:predicted} is shown in Fig.~\ref{fig:ndalignmentexample}.
Two of the variable names are exactly correct, but \cinline{buf} is more generic than \cinline{response} and \cinline{i} is a poor name for \cinline{retval}.
Only \approach can enable this type of variable name evaluation.

Table~\ref{tab:similarity} shows the results of several popular similarity metrics used to evaluate the same predictions by the same models.
The metrics agree that CodeT5+ performed better than CodeT5.
However, it's not clear by how much, and it's not clear how well each model performed in isolation in a way directly interpretable to humans.
These metrics are relative, rather than absolute, and offer substantially less detail.
Therefore, using these metrics allows weaker claims to be made about the performance of the models.

For instance Fig.~\ref{fig:original} and Fig.~\ref{fig:predicted} have a CodeBERTScore of 0.860.
It is difficult to know how good of a score this is, or to even understand why the score is what it is.
It is not the necessarily the case that an example given a higher score is better.
For example, these functions:
\begin{minipage}{0.5\textwidth}
\begin{lstlisting}[style=cstyle]
int main(void) {
    init();
    return auth() != 0;
}
\end{lstlisting}
\end{minipage}
\begin{minipage}{0.5\textwidth}
\begin{lstlisting}[style=cstyle]
int main(int argc, char *argv[]) {
    init();
    return auth() != 0;
}
\end{lstlisting}
\end{minipage}
are equivalent in C, but CodeBERTScore gives them a slightly lower score: 0.848. \approach aligns them perfectly.

\section{Discussion}
\label{sec:discussion}

\approach can be used to evaluate code generated as part of other machine learning tasks.
It works best when the generated code and reference code compute the result using the same algorithm.
Other machine learning tasks that have this property include transpilation and automated refactoring.
\approach will struggle in contexts where substantially different but semantically equivalent algorithms are acceptable. Of course, this is in general  undecidable.

\approach is also useful for tasks besides evaluating ML-generated code, like code clone detection 
(the parts of functions that align are clones).
Or, consider \emph{patch generation}, 
a critical part of automated program repair that creates code fragments to fix a bug in a piece of software.
An important subproblem in template-based patch generation is determining which variables should be instantiated inside a patch~\cite{tbar,sosrepair}.
An equivalence alignment can map the variables in the buggy region to those in a candidate patch in a similar way to how we mapped variable names to each other in Fig.~\ref{fig:ndalignmentexample}.
Third, 
\approach may also be useful for plagiarism detection.
Plagiarized code is by definition identical or very similar to the original.
\approach is robust to low-effort attempts to disguise plagiarism, such as variable renaming and (inconsequential) statement reordering, and the 
the equivalence alignment can be used as evidence.

\section{Related work}
\label{sec:relatedwork}

The most directly analagous piece of work to \approach is that of~\cite{yang1989detecting}.
They also operate on program dependence graphs and SSA-form code to determine if two functions are equivalent.
However, their technique is purely theoretical and handles an abstract, academic, feature-restricted language with only scalar variables and constants, and only if and while statements for control flow.
The last limitation means that instructions must have at most one control dependency.
Further, $\phi$ nodes must be definitively associated with either an `if' or `while' statement, limiting the complexity of the control flow to which their approach can scale.
In contrast, \approach has an implementation, can handle real C (\approach doesn't support \cinline{goto}s but theoretically could less irreducability) and an arbitrary number of control dependencies.

Another related line of work is \emph{semantic differencing}.
These tools output a \emph{functional} difference between two functions, such as a counterexample generated by an SMT solver.
Differential symbolic execution~\cite{DSE} identifies textual or AST-based differences in code, symbolically executes them, and queries the SMT solver to check their equivalence.
ArrDiff~\cite{ardiff} builds on this work by extracting information from unchanged code blocks that may nonetheless be useful for showing the diffs equivalent.
Symbolic-execution-based approaches are useful, but are unsound with respect to loops, are expensive to run, and must usually compile the input programs, which renders them useless for evaluating neural decompilers.
In contrast, SymDiff~\cite{symdiff} translates input functions into Boogie~\cite{boogie}, a verification language, and creates logical formulas summarizing the effects of the functions, then uses an SMT solver to check their equivalence.
This approach cannot fully handle loops; they must be unrolled to a specified depth or translated to tail-recursive functions, the latter of which are checked for equivalence separately.

Code clone detection~\cite{Roy09,Rattan13,zakeri2023systematic} tries to find functionally identical pieces of code so they can be refactored into a single entity for easier maintainability.
\approach can be used for code clone detection as discussed in Section~\ref{sec:discussion}.
Code clone techniques vary significantly based on the type of clones they attempt to target; \approach bears the strongest resemblance to program-dependence-graph (PDG)-based techniques~\cite{saha2013understanding}.
Isomorphic PDGs are code clones.
Unfortunately, finding maximal isomorphic subgraphs is NP-complete, so these techniques find approximations rather than true isomorphisms.
This renders these techniques unsound and unable to provide an equivalence alignment as \approach does.

Code similarity metrics~\cite{codebleu,codebertscore,crystalbleu,bleu,tran2019does} are used to evaluate code generated by machine learning models.
These are intended to measure how well a prediction matches the reference.
While the goal is to detect equivalent programs, these methods are necessarily heuristic and justified based on agreement with subjective human scores.
Unlike \approach, these measures offer a unitless similarity score which is meant to be compared with other values from the same metric.

Another related area is the formal, sound equivalence checking used to validate the correctness of optimizations produced by compilers.
A common abstraction used in this area is the \emph{simulation relation}, which bears some resemblance to an equivalence alignment.
A simulation relation is defined at program points, rather than for pairs of values, and contains symbolic relationships between live variables at those points.
The definition of a simulation relation does not provide a method for constructing the relation, and unlike an equivalence alignment, makes no suggestion as to what kinds relationships should be found at each program point.
Many translation validations techniques use compiler instrumentation to populate the simulation relation~\cite{necula2000translation,stepp2011equality,sewell2013translation}; this is not possible in neural decompilers because of their nature as machine learning models.
Likewise, approaches that use execution traces~\cite{churchill2019semantic} are not applicable to evaluating neural decompilers because decompiled code cannot, in general, be compiled.
The most directly applicable approaches are static, black-box translation validatiors, which makes no assumptions about the nature of the optimization performed.
In particular, work based on Joint Transfer Function Graphs (JTFGs), are the most analagous~\cite{dahiya2017black,gupta2018effective,gupta2020counterexample}, though in their current form, they require compilation (not execution).
These are simulation relations that bundle control flow with nonbranching code and attempt to match branches in the optimized and unoptimized with each other, usually with a heuristic guess-and-check strategy.
In contrast, \approach does not need heuristics to build an alignment; in particular, control flow alignment is efficient due to ordering.
JTFG approaches have substantial overhead, and take on the order of tens to hundreds of seconds per example even for very short functions of fewer than two dozen lines of code~\cite{gupta2018effective}. \approach is much faster.

\section{Conclusion}
\label{sec:conclusion}

In this work, we present \approach, a tool for evaluating neural decompilers.
\approach generates and abstraction called an equivalence alignment, which states which parts of the input functions are equivalent.
We show \approach can be used to evaluate the correctness of neural decompilers' predictions.
Further, we show how an equivalence alignment can be used to determine which variable names in the decompiled code map to those in the reference, allowing us to evaluate variable name quality as well.
We discuss how the equivalence alignment abstraction has applications to a variety of different tasks, including program repair, code clone detection, and plagiarism detection.

\bibliographystyle{ACM-Reference-Format}
\bibliography{references}

\end{document}